\def\year{2019}\relax
\documentclass[letterpaper]{article} 
\usepackage{aaai19}  %Required
\usepackage{times}  %Required
\usepackage{helvet}  %Required
\usepackage{courier}  %Required
\usepackage{url}  %Required
\usepackage{graphicx}  %Required
\usepackage{subfigure}
\usepackage{comment}

\usepackage{algorithm}
\usepackage{algorithmic}

\usepackage{amsmath}
\usepackage{amssymb}

\usepackage{url}

\nocopyright

\title{Soft Label Memorization-Generalization for Natural Language Inference}

\author{John P. Lalor$^{1}$, Hao Wu$^{2}$, Hong Yu$^{1,3}$\\
$^1$ University of Massachusetts, %Amherst, MA, 
$^2$ Vanderbilt University, %Chestnut Hill, MA, 
$^3$ Bedford VA\\
\tt{lalor@cs.umass.edu}, \tt{hao.wu.1@vanderbilt.edu}, \tt{hong.yu@umassmed.edu}}

\begin{document}
\maketitle

\begin{abstract}

\begin{comment}
Human language is ambiguous, however ambiguity is often not represented in most data sets used in machine learning. 
For single-label classification tasks The standard assumptions is that each example has a single correct label, and each example for a label is equally useful for classifying that label.
In reality, some training examples may be more ambiguous than others. 
As a result, there maybe a distribution of human-judged labels.  
In this work we propose new methods and data that leverage crowdsourced labels to model ambiguity as a distribution over labels.
By fine-tuning ML models with the distribution over labels instead of a single gold-standard label, we report improved performance for two NLP tasks: Recognizing Textual Entailment (NLI) and Sentiment Analysis (SA).
Models trained with the crowd-informed fine-tuning data classify more instances of entailment and contradiction labels correctly for the NLI task at the expense of neutral examples. 
\end{comment}
Often when multiple labels are obtained for a training example it is assumed that there is an element of noise that must be accounted for.
It has been shown that this disagreement can be considered signal instead of noise.
In this work we investigate using soft labels for training data to improve generalization in machine learning models.
However, using soft labels for training Deep Neural Networks (DNNs) is not practical due to the costs involved in obtaining multiple labels for large data sets.
We propose soft label memorization-generalization (SLMG), a fine-tuning approach to using soft labels for training DNNs.
We assume that differences in labels provided by human annotators represent ambiguity about the true label instead of noise.
Experiments with SLMG demonstrate improved generalization performance on the Natural Language Inference (NLI) task.
Our experiments show that by injecting a small percentage of soft label training data (0.03\% of training set size) we can improve generalization performance over several baselines.
%\end{comment}

\end{abstract}

\section{Introduction}
\label{sec:introduction}

In Machine Learning (ML) classification tasks a model is trained on a set of labeled data and optimized based on some loss function.
The training data consists of some feature set $X_{\text{train}} = {x_i,\dots, x_N}$ and associated labels $Y_{\text{train}} = {y_1,\dots,y_N}$, where $Y$ is a vector of integers corresponding to the classes of the problem.
%For binary classification, $Y$ would be a vector of $0$'s and $1$'s, with $0$ representing the negative class and $1$ representing the positive class.
%The goal when training an ML classification model is to minimize the error in the model's prediction of a class label for a given training example.
%The loss function can take many forms, but at a high level we want to minimize the number of training examples the model misclassifies: $\sum_i^N \mathbbm{1} [\hat{y_i} \neq y_i ]$ where $\mathbbm{1}[x]$ is the indicator function.
Typically we assume that each training example is labeled correctly, and each is equally appropriate for a single class.
There is no way to quantify the uncertainty of the examples, nor a way to exploit such uncertainty during training.
Particularly for NLP tasks with sentence- or phrase-based classification such as Natural Language Inference (NLI), it is not common to model ambiguity in language in training data labels.

For example, consider the following two premise-hypothesis pairs, both taken from the Stanford Natural Language Inference (SNLI) corpus for NLI~\cite{bowman_large_2015}:

\begin{enumerate}
	\item \textit{Premise:} Two men and a woman are inspecting the front tire of a bicycle. \\ \textit{Hypothesis:} There are a group of people near a bike.
	\item \textit{Premise:} A young boy in a beige jacket laughs as he reaches for a teal balloon.\\ \textit{Hypothesis:} The boy plays with the balloon.
\end{enumerate}

In both cases the gold-standard label in the SNLI data set is \textit{entailment}, which is to say that if we assume that the premise is true, one can infer that the hypothesis is also true.
However, looking at the two sentence pairs one could argue that they do not both equally describe entailment.
The first example is a clear case: people inspecting a front tire of a bike are almost certainly standing near it.
However the second example is less clear. 
Is the child laughing because he is playing? 
Or is he laughing for some other reason, and is simply grabbing for the balloon to hold it (or give it to someone else)?
There is ambiguity associated with the two examples that is not captured in the data.
To a machine learning model trained on SNLI, both examples are to be classified as entailment, and incorrect classifications should be penalized equally during learning.

Previous work has shown that leveraging crowd disagreements can improve the performance of named entity recognition (NER) models by treating disagreement not as noise but as signal~\cite{inel2017harnessing}.
We use the same assumption here and encode crowd disagreements directly into the model training data in the form of a distribution over labels (``soft labels'').
These soft labels model \textit{uncertainty} in training by representing human \textit{ambiguity} in the class labels.
Ideally we would have soft labels for all of our training data, however when training large deep learning models it is prohibitively expensive to collect many annotations for all data in the huge datasets required for training.
In this work we show that even a small amount of soft labeled data can improve generalization.
This is the first work to fine-tune a deep neural network with soft labels from crowd annotations for a natural language processing (NLP) task.

With this in mind we propose soft label memorization-generalization (SLMG), a fine-tuning approach to training that uses distributions over labels for a subset of data as a supplemental training set for a learning model.
Ideally a model could be trained with soft labels for all training examples, but because of the costs involved, in this work we explore using a small number of examples for fine-tuning on top of a larger data set.
We seek understand the effect of including more informative labels as part of training.

Our hypothesis is that using labels that incorporate language ambiguity can improve model generalization in terms of test set accuracy, even for a small subset of the training data. 
By using a distribution over labels we hope to reduce overfitting by not pushing probabilities to $1$ for items where the empirical distribution is more spread out. 
Our results show that SLMG is a simple and effective way to improve generalization without a lot of additional data for training.

We evaluate our approach on NLI (also known as Recognizing Textual Entailment or RTE)~\cite{dagan_pascal_2006} using the SNLI data set~\cite{bowman_large_2015}.
Prior work has shown that lexical phenomena in the SNLI dataset can be exploited by classifiers without learning the task, and performance on difficult examples in the data set is still relatively poor, making NLI a still-open problem \cite{gururangan2018annotation,poliak2018hypothesis,lalor2018irt}.
For soft labeled data we use the IRT evaluation scales for NLI data~\cite{lalor2016beyond} where each premise-hypothesis pair was labeled by 1000 AMT workers.
This way we are able to leverage an existing source of soft labeled data without additional annotation costs.
We find that SLMG can improve generalization under certain circumstances, even thought the amount of soft labeled data used is tiny compared to the total training sets (0.03\% of the SNLI training data set).
SLMG outperforms the obvious but strong baseline of simply gathering more unseen data for labeling and training.
Our results suggest that there are diminishing returns for simply adding more data past a certain point~\cite{halevy2009unreasonable}, and indicate that representing data uncertainty in the form of soft labels can have a positive impact on model generalization.

Our contributions are as follows: (i) We propose the SLMG framework for incorporating soft labels in machine learning training, (ii) We use previously-collected human annotated data to estimate soft label distributions for NLI and show that replacing less than 0.1\% of training data with soft labeled data can improve generalization for three DNN models, and (iii) We demonstrate for the first time that soft labels can encode ambiguity in training data that can improve model generalization in terms of test set accuracy.\footnote{We will release our code upon publication.}

\begin{comment}
(i) we introduce a fine-tuning method for incorporating crowd uncertainty into training, (ii) we demonstrate a new use case for the NLI IRT data set \cite{lalor2016beyond} and release a new data set of AMT responses for a subset of SSTB \cite{socher2013recursive}\footnote{Data included as supplemental material. We will release our code upon publication.}, (iii) we conduct experiments to show where the fine-tuning data is useful in a model training setup, and (iv) we analyze changes in test set output to understand what types of examples are impacted by SLMG.
% TODO: redo
The rest of this paper is organized as follows: Section \ref{sec:SLMG} motivates the need for uncertainty in training labels and describe SLMG, Section \ref{sec:background} gives a thorough description of related work to place SLMG in context, Section \ref{sec:experiments} describes our experiments, including the deep learning models we tested, Section \ref{sec:results} presents results, and Section \ref{sec:discussion} discusses the results and areas for potential future work.
\end{comment}
\section{Soft Label Memorization-Generalization}
\label{sec:slmg}
\subsection{Overview}
\label{ssec:slmg_intro}

In a traditional supervised learning single-label classification problem, a model is trained on some data set $X_{\text{train}}$, and tested on some test set $X_{\text{test}}$.
In this setting, learning is done by minimizing some loss function $L$. 
We assume that the labels associated with instances in $X_{\text{train}}$ are correct.
That is, for each $(x_i, y_i) \in X_{\text{train}}$ we assume that $y_i$ is the \textit{correct} class for the $i$-th example, where $x_i$ is some set of features associated with the $i$-th training example and $y_i$ is the corresponding class.
However it is often the case, particularly in NLP, that examples may vary in terms of difficulty, ambiguity, and other characteristics that are often not captured by the single correct class to which the example belongs.
The traditional single-label classification task does not take this into account.

For example, a popular loss function for classification tasks is Categorical Cross-Entropy (CCE).
For a single training example $x_i$ with class $y_i \in Y$ where $Y$ is the set of possible classes, CCE loss is defined as $L_i^{\text{CCE}} = -\sum_{j=1}^{\vert Y \vert} p(y_{ij})\log p(\hat{y}_{ij})$.
In the single-class classification case where a single class $j$ has probability $1$ CCE loss is $L^{\text{CCE}} = \sum_i^N -\log p(\hat{y}_{ij})$, where each example loss is summed over all of the training examples.
With this loss function a learning model is encouraged to update its parameters in order to maximize the probability of the correct class for each training example.
Without some stopping criteria, parameter updates will continue for a given example until $p(\hat{y}_{ij}) = 1$.
This may not always be ideal, since by pushing the model output probability to $1$, the learner is encouraged to overfit on an example that may not be representative of the particular class.

\begin{table*}[th]
	\small
	\centering
	
	\begin{tabular}{p{6cm}p{6cm}lll}
		\bf Premise & \bf Hypothesis & \bf $P(\text{E})$ & \bf $P(\text{C})$ & \bf $P(\text{N})$\\ \hline
		A little boy is opening gifts surrounded by a group of children and adults. & The boy is being punished&0.005&\bf 0.839&0.156 \\
		A man and woman walking away from a crowded street fair.& There are a group of men walking together.& 0.045&\bf 0.542& 0.412\\
		Two men and a woman are inspecting the front tire of a bicycle.&There are a group of people near a bike. &\bf 0.861&0.032&0.108\\
		A young boy in a beige jacket laughs as he reaches for a teal balloon.& The boy plays with the balloon.&\bf 0.659&0.026&0.316\\ 
		A man wearing a gray shirt waving in the middle of a plant nursery&The man does not have a way to get home.&0.011&0.174&\bf 0.815 \\
		A wielder works on wielding a beam into place while other workers set beams.&The wielder is working on a building.&0.486&0.013&\bf 0.501\\ \hline
	\end{tabular}
	\caption{Examples of premise-hypothesis pairs from the SNLI data set and the AMT-estimated probability that the correct label is Entailment (\textbf{E}), Contradiction (\textbf{C}), or Neutral (\textbf{N}). The original gold-standard label from SNLI is bolded. In some cases, the gold label provided originally has a low probability based on AMT-population estimates (i.e. less than 75\%).}
	\label{tab:examples}
\end{table*}

\begin{comment}
\begin{table*}[th]
	\scriptsize
	\centering
	
	\begin{tabular}{p{8.0cm}ccccc}
		\bf Sentence & \bf $P(\text{VNeg})$ & \bf $P(\text{Neg})$ & \bf $P(\text{Neu})$ & \bf $P(\text{Pos})$ & \bf $P(\text{VPos})$\\ \hline
		An endlessly fascinating, landmark movie that is as bold as anything the cinema has seen in years.&0.01 &0.015 &0.023&0.128&\bf 0.824 \\
		If no one singles out any of these performances as award-worthy, it's only because we would expect nothing less from this bunch.&0.061 &0.148 &0.164&0.297&\bf 0.329 \\
		Trivial where it should be profound, and hyper-cliched where it should be sincere&\bf 0.421&0.416&0.093&0.048&0.021\\ \hline
	\end{tabular}
	\caption{Examples from the SSTB data set and the AMT-estimated probabilities over labels. The gold label from SSTB is bolded.}
	\label{tab:examples_sentiment}
\end{table*}
\end{comment}

With SLMG we want to take advantage of the fact that differences between examples in the same class can be useful during training.
Instead of treating each training example as having a single correct class, SLMG uses a distribution over labels for the gold standard.
This way examples with varying degrees of uncertainty are reflected during training.

We make a different assumption regarding noise in human generated labels than previous work~\cite{dawid1979maximum,ICML2012Bachrach_597}.
The presence of noise when multiple labels are obtained is often attributed to labeler error, lack of expertise, adversarial actions, or other negative causes.
However, we believe that the noise in the labels can be considered a \textit{signal}~\cite{inel2014crowdtruth,aroyo2015truth}.
Examples with less uncertainty about the label (in the form of a label distribution with a single high peak) should be associated with similarly high model confidence.

\subsection{Training with SLMG}

In our experiments we investigated two ways to incorporate the soft labeled data into model training, which we define below.
Let $X_{\text{train}}$ be the \textit{original} training set, and let $X_{\text{test}}$ be the test set.
Let $X_{\text{soft}}$ be the soft labeled training data with class probabilities.
There are two ways to incorporate the $X_{\text{soft}}$ data into a learning task that we investigate: (i) at each training epoch, training with $X_{\text{train}}$ and $X_{\text{soft}}$ \textit{interspersed} (SLMG-I), and (ii) train a model on $X_{\text{train}}$ for a predefined number of epochs, followed by training on $X_{\text{soft}}$ for a predefined number of epochs, repeated some number of times (\textit{meta-epochs}) in a \textit{sequential} fashion (SLMG-S). 
Algorithms \ref{alg:SLMG_int} and \ref{alg:SLMG_seq} define the two training sequences, respectively.
In our experiments we tested two loss functions for the SLMG data, CCE (\S \ref{ssec:slmg_intro}) and Mean Squared Error (MSE): $L^{\text{MSE}}_i =  \sum_{j=1}^{\vert Y \vert} (\hat{p}(y_{ij}) - p(y_{ij}))^2$.

\subsubsection{Interspersed Fine-Tuning}
The motivation for interspersing fine-tuning with soft labels is to prevent overfitting as the model learns. 
After each epoch in the training cycle, the learning model will have made updates to the model weights according to the outputs on the full training set. 
By interspersing the fine-tuning after each epoch, our expectation is that we can account for and correct overfitting earlier in the process by making smaller updates to the model weights according to the soft label distributions.
This method encourages generalization early in the process, before the model can memorize the training data and possibly overfit.

\subsubsection{Sequential Fine-Tuning} 
In contrast with the interspersed fine-tuning, the motivation for sequential fine-tuning is to adjust a well-trained model to improve generalization.
After a full training cycle of some number of epochs, the learning model is then fine-tuned using the soft-labeled data.
This way the fine-tuning takes place after the model has learned a set of weights that perform well on the training data.
Fine-tuning here can improve generalization by updating the model weights to be less extreme when dealing with examples that are more ambiguous than others. 
Since these updates happen on a trained model, there is less risk of the model performance drastically reducing. 
By repeating this process over a number of meta-epochs, the learning model can memorize, generalize, and repeat the cycle.

\begin{algorithm}[t!]
	\caption{SLMG-I Algorithm}
	\label{alg:SLMG_int}
	\begin{algorithmic}
		\STATE {\bfseries Input:} Model $m$, NumEpochs $e$, $X_{\text{train}}$, $X_{\text{soft}}$
		\FOR{$i=1$ {\bfseries to } $e$}
		\STATE Train $m$ on $X_{\text{train}}^N$
		\STATE Train $m$ on $\text{X}_{\text{soft}}$ 
		\ENDFOR
	\end{algorithmic}
\end{algorithm}

\begin{algorithm}[t!]
	\caption{SLMG-S Algorithm}
	\label{alg:SLMG_seq}
	\begin{algorithmic}
		\STATE {\bfseries Input:} Model $m$, NumMetaEpochs $me$, NumEpochs $e$, $X_{\text{train}}$, $X_{\text{soft}}$
		\FOR{$i=1$ {\bfseries to} $me$}
		\FOR{$j=1$ {\bfseries to } $e$}
		\STATE Train $m$ on $X_{\text{train}}^N$
		\ENDFOR
		\FOR{$j=1$ {\bfseries to } e}
		\STATE Train $m$ on $\text{X}_{\text{soft}}$ 
		\ENDFOR
		\ENDFOR
	\end{algorithmic}
\end{algorithm}

\subsection{Collecting Soft Labeled Data}

For our NLI soft labeled data, we use data collected by~\cite{lalor2016beyond}.
180 SNLI training examples split evenly between the three labels were randomly selected and given to Amazon Mechanical Turk (AMT) workers (Turkers) for additional labeling.
For each example 1000 additional labels were collected.
%In order to generate a distribution over labels for these examples, we use the Turker responses to estimate the probability of each label based on the number of labels for each class.
In order to estimate a distribution over labels for these examples we calculate the probability of a certain label according to the proportion of humans that selected the label: $P(Y = y) = \frac{N_y}{N}$, where $N_y$ is the number of times $y$ was selected by the crowd and $N$ is the total number of responses obtained.

\begin{comment}
For SA, we collected a new data set of labels for 134 examples randomly selected from SSTB, using a similar AMT setup as \cite{lalor2016beyond}.
For each randomly selected example, we asked 1000 Turkers to label the sentence as very negative, negative, neutral, positive, or very positive. 
\end{comment}

Table \ref{tab:examples} shows example premise-hypothesis pairs taken from the SNLI data set for NLI~\cite{bowman_large_2015}.
Table \ref{tab:examples} includes the premise and hypothesis sentences, the gold standard class as included in the data set, as well as estimated soft labels using human responses obtained by~\cite{lalor2016beyond}.
There are premise-hypothesis pairs that share a class label (e.g. the first two examples) yet are very different in terms of how they are perceived by a crowd of human labelers.
In a traditional setup both examples would have a single class label associated with contradiction (class label $1$ if $0$ = \textit{entailment}, $1$ = \textit{contradiction}, and $2$ = \textit{neutral}).
Certain training examples have much less uncertainty associated with them, which is reflected in the high probability weight on the correct label.
In other cases, there is a more evenly spread distribution, which can be interpreted as a higher degree of uncertainty.
In a learning scenario, one may want to treat these examples differently according to their uncertainty, as opposed to the common practice of weighing each equally.
%Similarly in Table \ref{tab:examples_sentiment} we see examples of sentences for the Sentiment Analysis (SA) task that demonstrate how the gold standard label provided in the data set does not capture the uncertainty with the labels provided by the crowd.

Consider calculating the entropy, $H(X)$, of the first two training examples from Table \ref{tab:examples}: $H(X) = - \sum_{y \in Y} p(y) \log p(y)$.
If we assume that the probability of the correct label (in this case, contradiction), is $1$, and the probability of all other labels is $0$, then entropy in both cases is 0.\footnote{Where $0 \log 0 = 0$.}
However if we use the distributions from Table \ref{tab:examples}, then entropy is $0.464$ and $0.837$ respectively.
There is much more uncertainty in the second example than the first, which is not reflected if we assume that both examples are labeled contradiction with probability $1$.
This uncertainty may be important when learning for classification.

\subsection{Learning from the Crowd}
\label{ssec:crowd}

In this work we take advantage of the fact that we have a distribution over labels provided by the human labelers.
We can train using CCE or MSE: 
as our loss function, where we minimize the difference between the estimated probabilities learned by the model and the empirical distributions obtained from AMT over the training examples.
With SLMG we are attempting to move the model predictions closer to the soft label distribution of responses.
We are not necessarily trying to push predicted probability values to 1, which is a departure from the standard understanding of single label classification in ML.
Here we hypothesize that updating weights according to differences in the observed probability distributions will improve the model by preventing it from updating too much for more uncertain items (that is, examples where the empirical distribution is more evenly spread across the three labels).

This scenario assumes that the crowdsourced distribution of responses is a better measure of correctness than a single gold-standard label.
We hypothesize that the crowd distribution over labels gives a fuller understanding of the items being used for training.
SLMG can update parameters to move closer to this distribution without making large parameter updates under the assumption that a single correct label should have probability 1.

If we assume that ML performance is not at the level of an average human (which is reasonable in many cases), then SLMG can help pull models towards average human behavior when we use human annotations to generate the soft labels.
If the model updates parameters to minimize the difference between predictions and the distribution of responses provided by AMT workers, then the model predictions should look like that of the crowd.
When ML model performance is better than the average AMT user, there is a risk that performance may suffer, if we assume that our model would outperform a human population.
The model may have learned a set of parameters that better models the data than the human population, and updating parameters to reflect the human distribution could lead to a drop in performance.
However since we are only using SLMG as a fine-tuning mechanism, the risk here is mitigated by the larger training set that we use alongside the SLMG data.

\section{Experiments}
\label{sec:experiments}

Our hypothesis is that soft labeled data, even in very small amounts, can improve model generalization by capturing ambiguity of language data in the form of distributions over labels.
In this section we describe our experiments to test this hypothesis, as well as the data sets and models used in the experiments.
%At a high level, our goal is to understand how distributions over labels can affect the learning process.
%To do this we look at several ways of incorporating the SLMG data.
%By varying the point at which we inject the SLMG data we can observe how performance is affected.

\subsection{Models}
\label{ssec:models}

For our experiments we tested three deep learning models, an LSTM RNN~\cite{hochreiter_long_1997,bowman_large_2015} that was released with the original SNLI data set, a memory-augmented LSTM network~\cite{munkhdalai2016neural}, and a recently released hierarchical network with very strong performance on the SNLI task~\cite{Chen-Qian:2017:ACL}.
Each model was trained according to the original parameters provided in the respective papers.\footnote{Due to space constraints, please refer to the original papers for descriptions of the model architectures.}
Word embeddings for all models were initialized with GloVe 840B 300D word embeddings~\cite{pennington2014glove}.

%\subsubsection{LSTM}

Our first model is a re-implementation of the 100D LSTM model that was released with the original SNLI data set~\cite{bowman_large_2015}.
For the NLI task, the premise and hypothesis sentences were both passed through a 100D LSTM sequence embedding~\cite{hochreiter_long_1997}.
The output embeddings were concatenated and fed through 3 200D tanh layers, followed by a final softmax layer for classification.
%For SA, the concatenation step was removed, and the tanh layers were 100D, but the rest of the process was the same. 
We implemented in DyNet~\cite{dynet}.

%\subsubsection{Neural Semantic Encoders}

Neural Semantic Encoder (NSE)~\cite{munkhdalai2016neural} is a memory augmented neural network.
NSE uses read, compute, and write operations to maintain and update an external memory $M$ during training and outputs an encoding $h$ that is used for downstream classification tasks:
\begin{comment}
\begin{align*}
o_t &= f_{\text{r}}^{LSTM}(x^t) \\
z_t &= softmax(o_t^{\top}M_{t-1}) \\
m_{r,t} &= z_t^{\top}M_{t-1} \\
c_t &= f_{\text{c}}^{MLP}(o_t,m_{r,t}) \\
h_t &= f_{\text{w}}^{LSTM}(c_t) \\
\begin{split}
M_t = M_{t-1}(&\mathbf{1} - (z_t \otimes e_k)^{\top}) + (h_t \otimes e_t)(z_t \otimes e_k)^{\top}
\end{split}
\end{align*}
where $f_r^{LSTM}$ is the read function, $f_c^{MLP}$ is the composition function, $f_w^{LSTM}$ is the write function, $M_t$ is the external memory at time $t$, and $e_l \in R^l$ and $e_k \in R^k$ are vectors of ones~\cite{munkhdalai2016neural}.
\end{comment}
%
We used the publicly available version of the NSE model released by the authors\footnote{\url{https://bitbucket.org/tsendeemts/nse}} and implemented in Chainer~\cite{chainer_learningsys2015}.
We followed the original NSE training parameters and hyperparameters~\cite{munkhdalai2016neural}.

%\subsubsection{Enhanced Sequential Inference Model}

The Enhanced Sequential Inference Model (ESIM)~\cite{Chen-Qian:2017:ACL} consists of three stages: (i) input premise and hypothesis encoding with BiLSTMs, (ii) local inference modeling with attention, and (iii) inference composition with a second BiLSTM encoding over the local inference information.
We used the publicly available ESIM model released by the authors\footnote{\url{https://github.com/lukecq1231/nli}} implemented in Theano~\cite{2016arXiv160502688short} and kept all of the hyperparameters the same as in the original paper.
%We do not use ESIM in our SA experiments.
%The model was designed specifically for NLI, as opposed to NSE which performs well across several tasks including sentiment analysis \cite{munkhdalai2016neural}.
%In addition, NSE performance on SA is close to state-of-the-art, so testing another high-performing model in this case is unnecessary.

\begin{comment}
\begin{table*}[t]
	\small
	\centering
	
	\begin{tabular}{ll|l|llll}
		\bf Task & \bf Model & \bf Baseline &  \multicolumn{2}{c}{\bf $\text{SLMG-S}$} & \multicolumn{2}{c}{\bf $\text{SLMG-I}$} \\ \hline
		&&&MSE & CCE & MSE & CCE  \\ \hline 
		&LSTM & 76.7 & 76.5 & \bf 77.4 & 76.9 & 76.7 \\
		NLI &NSE & 84.6 & 84.1 & \bf 85.1 & 84.3 & 84.4 \\
		&ESIM & 87.7 & 87.7 & 87.6 & 87.8 &\bf 87.9  \\ \hline
		SA-B&LSTM & 87.4  & 87.3 & \bf 87.5 & 86.5  & \bf 87.5 \\
		&NSE & 88.9 & 87.5 & 88.6 & 88.4 & \bf 89.1 \\ \hline
		SA-FG&LSTM & 49.7 & 50.8 & 47.0 & \bf 51.7  & 49.9 \\
		&NSE & \bf 52.3 & 51.0 & 51.9 & 52.0 & 51.2 \\
		\hline
	\end{tabular}
	\caption{Training and test accuracy results for incorporating SLMG in three tasks: NLI, binary sentiment analysis (SA-B), and fine-grain sentiment analysis (SA-FG).}
	\label{tab:results}
\end{table*}
\end{comment}

\subsection{Data}

For NLI data we used the SNLI corpus~\cite{bowman_large_2015}.
SNLI is an order of magnitude larger than previously available NLI data sets (550k train/10k dev/10k test), and consists entirely of human-generated P-H pairs.
SNLI is evenly split across three labels: entailment, contradiction, and neutral.
%For SA, our main data set is the Stanford Sentiment Treebank (SSTB), which includes sentence- and phrase-level sentiment labels for 11,000 sentences (215,000 phrases).
%SSTB includes binary (positive/negative) and fine-grained (very negative/negative/neutral/positive/very positive) labels.
SNLI is large, well-studied, and often used as a benchmark for new NLP models for NLI.

\begin{comment}
The SA soft label data examples were selected from the SSTB test set, so for our experiments we use a modified SSTB test set where the examples have been removed.
In our results we report baseline scores on the modified test set so as to be consistent.
We chose to select from the SSTB test set because the training set for SSTB, particularly for the binary task, is smaller than the SNLI data set.
We would rather keep all data for training in this instance, and report all of our results on a smaller, but still substantial test set.
For all experiments we used early stopping and report test results for the epoch with the highest dev set performance.
\end{comment}

\begin{table*}[th!]
	\small
	\centering
	
	\begin{tabular}{p{4.5cm}p{4.5cm}llll}
		\bf Premise & \bf Hypothesis & \bf Model& \bf $P(\text{E})$ & \bf $P(\text{C})$ & \bf $P(\text{N})$\\ \hline
		This church choir sings to the masses as they sing joyous songs from the book at a church.&The church is filled with song&B1&\bf 0.191& 0.021&0.788 \\
		& & SLMG-I-CCE&\bf 0.520&0.028&0.452 \\ \hline 
		A land rover is being driven across a river.&A sedan is stuck in the middle of a river.& B1&0.014&\bf0.561 &0.435 \\ 
		& & SLMG-I-CCE&0.011 & \bf 0.241 &0.749  \\ \hline
	\end{tabular}
	\caption{Examples of premise-hypothesis pairs from the SNLI data set and output probabilities from the LSTM model. For both examples the probabilities associated with the gold label are bolded.}
	\label{tab:model_outputs}
\end{table*}

\begin{table}[t]
	\small
	\centering
	
	\begin{tabular}{l|ccc}
		\bf Experiment & \multicolumn{3}{c}{\bf Model} \\
		\bf &\bf LSTM & \bf NSE &\bf ESIM \\ \hline 
		B1: Traditional & 76.7& 84.6& 87.7\\
		B2: CLE & 76.9&84.8 &87.1 \\
		B3: AOC & 75.7& 84.0& 87.7\\
		SLMG-S-MSE & 76.5& 84.1& 87.7\\
		SLMG-S-CCE & \bf 77.4& \bf 85.1& 87.6\\
		SLMG-I-MSE & 76.9& 84.3& 87.8\\
		SLMG-I-CCE & 76.7& 84.4& \bf 87.9\\ \hline
		
	\end{tabular}
	\caption{Test accuracy results for incorporating SLMG for NLI. Refer to \S \ref{ssec:baselines} for descriptions of the baselines. Highest accuracy result for each model is bolded (one per column).}
	\label{tab:results}
\end{table}

\subsection{Baselines}
\label{ssec:baselines}

We evaluate SLMG against three baselines: (i) \textit{B1, Traditional:} We train the DNN models (\S \ref{ssec:models}) in a traditional supervised learning setup, where the soft labeled training data ($X_{soft}$) is incorporated in the hard labeled training data ($X_{train}$) with their original gold-standard labels, (ii) \textit{B2, Comparable Label Effort (CLE):} Because each of the 180 $X_{soft}$ examples have 1000 human annotations, our second baseline is to add new single label training data to B1, to evaluate against a comparable data labeling effort. To that end, we randomly selected 180,000 additional training data points from the Multi-NLI data set~\cite{williams_multinli_2018} for additional training data, (iii) \textit{B3, AOC:} The third baseline is the All in one Classifier (AOC) approach proposed by~\cite{kajino2012convex}, where for each example in $X_{soft}$, every label obtained from the crowd is used as a unique example in the training data. This baseline also has an addition 180,000 training data points as in B2, but the additional pairs all come from $X_{soft}$ and have varying labels depending on the crowd responses.

\section{Results and Analysis}
\label{sec:results}

Table \ref{tab:results} reports results on the SNLI test set.
For each model on the NLI task, we are able to improve generalization performance (i.e. test set accuracy) by injecting soft labeled data at some point.
% include the following sentence if true once ESIM experiment is complete.
%In fact by using SLMG for the ESIM model, we have SNLI test results that improve over the latest state-of-the-art results \cite{Chen-Qian:2017:ACL}.
Note that the best performance with SLMG varies according to the model, but for each model there is some configuration that does improve performance.
As with all model training, the effect of SLMG requires experimentation according to the use case.
In all cases, using CCE as the loss function performs better than using MSE. 
We suspect that this is due to the fact that small differences are penalized less with CCE than with MSE.

Table \ref{tab:model_outputs} shows example of two premise-hypothesis pairs from the SNLI test set, and the model output probabilities from the B1 baseline and the SLMG-I model trained with CCE as the soft label loss function.
In the first example, using SLMG results in flipping the output from incorrect (neutral) to correct (entailment).
However, this pair seems to be a weak case of entailment, and could be argued to be neutral.
The SLMG model considers this and has a reasonably high probability for the neutral class.
In the second case, training with SLMG results in the wrong label, but again it could be argued that this is a case where neutral is appropriate.
The ``sedan'' that is stuck may not be the Land Rover (Land Rovers are SUVs), so neutral is a reasonable output here.

\begin{comment}
For the SA task, injecting SLMG data at some point again improves performance.
SLMG does not improve performance for the NSE model on the fine-grained SA task, but for the binary task we do see improvements, both for the LSTM and NSE model.
This suggests that data close to the decision boundary that was originally misclassified was classified correctly when soft labeled data was added. 
With binary SA, there is no distinction between ``very negative'' and ``negative'' so changes in degree don't have an effect, unless the change is from negative to positive.
\end{comment}

\subsection{Changes in Outputs from SLMG}
\begin{table}[t!]
	\small
	\centering
	\begin{tabular}{ll|lll}	
		&& E & C & N \\ \hline
		&E & 2739 & 191 & 438 \\
		\bf Baseline &C &333&2360&544\\
		&N &441&332&\bf 2446 \\ \hline
		&E &2828 &157 &383 \\
		\bf SLMG-S (CCE)&C &375&2401&461\\
		&N &520&328&2371 \\ \hline
		&E &\bf 2967 &158 &243 \\
		\bf SLMG-S (MSE)&C &466&\bf 2415&356\\
		&N &677&422&2120 \\ \hline
	\end{tabular}
	\caption{Confusion matrices for the LSTM model, trained according to the baseline (first block), using SLMG-S with CCE  (second block), and using SLMG-S with MSE (third block). Gold standard labels run down the left hand side, while predicted labels are across the top in the matrix. The highest count of True Positives for each label across the three model-training setups are bolded.}
	\label{tab:confmatrices}
\end{table}
To better understand the effects of SLMG on generalization, we look at the changes in test set performance when SLMG is used as compared to the baseline case.
Table \ref{tab:confmatrices} shows 3 confusion matrices: the test-set output for the baseline LSTM model on the NLI task, and the same model when trained with $\text{SLMG-S}$ and CCE as the loss function for the soft labeled data, which improved test set performance and $\text{SLMG-S}$ with MSE as the loss function for the soft labeled data, which did not.
In both cases of training with SLMG, the number of correctly classified entailment and contradiction examples increased, while the number of neutral examples correctly classified decreased.
However when MSE is used as the soft label loss function, the increase in misclassified neutral examples was enough to offset the gains in correctly classified entailment and contradiction examples.
Depending on the use case, this result could be useful for applications.
Fewer false negatives for entailment and contradiction examples may be more important than fewer true positives for the neutral class.

If we consider SNLI as a binary classification task, with two possible labels ``entailment'' and ``not entailment'' (where we combine contradiction and neutral), and look at Table \ref{tab:confmatrices} we see that SLMG outperforms the baseline in both cases.
In fact, the SLMG-MSE method outperforms SLMG-CCE in the binary task ($88.0\%$ vs. $86.6\%$) due to the fact that its performance on the entailment label is much higher.

\subsection{Comparing the Crowd to the Gold Standard}

\begin{figure}[t!]
	\centering
	\includegraphics[width=0.8\columnwidth]{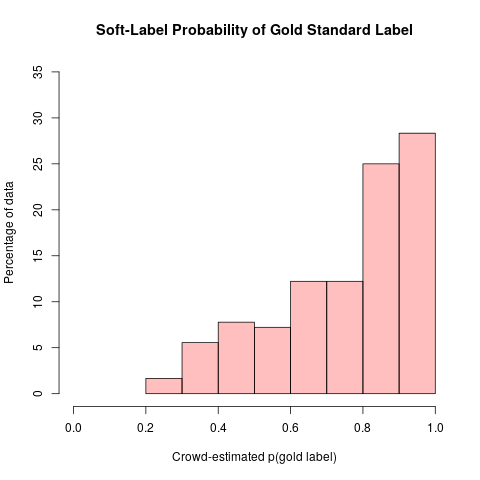}
	\caption{Relative frequency histograms for the crowd-estimated probability of the original gold-standard label. 
	}
	\label{fig:histogram}
\end{figure}

We also looked at the soft labeled data itself to understand how well the crowd label distributions align with the accepted gold-standard labels in the original data set.
Figure \ref{fig:histogram} reports on how well the crowd distributions align with the gold standard labels included in the original SNLI data set.
We see that there are quite a few examples where the gold standard class label does not have a high degree of probability weight as estimated from the crowd.

For NLI, there is a high percentage of examples where the gold label has an estimated probability of less than $80\%$. 
This may be due to the fact that individuals have different understanding of what constitutes entailment.
This uncertainty among humans is useful for understanding outputs from ML models.
This is consistent with the inter-rater reliability (IRR) scores originally reported by~\cite{lalor2016beyond} with the IRT data set. 
IRR scores (Fleiss' $\kappa$) for the data ranged from $0.37$ to $0.63$, which is considered moderate agreement~\cite{landis1977measurement}.
The moderate agreement indicates that there is a general consensus about which label is correct (which is consistent with Figure \ref{fig:histogram}), but there is enough disagreement among the annotators that the disagreements should be incorporated into the training data, and not discarded in favor of majority vote or another single label selection criteria.
\begin{figure}[t!]
	\centering
	\includegraphics[width=0.8\columnwidth]{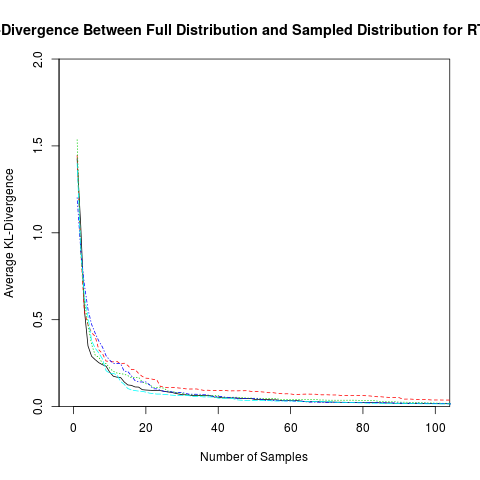}
	\caption{Average KL-Divergence between sub-sampled crowd distributions and the estimated soft label distribution from the entire crowd data. By sampling 20 crowd workers we achieve a good estimate of the label distributions without the cost of using the full 1000 worker population.}
	\label{fig:kld}
\end{figure}

\subsection{How Many Labels do we Need?}

Of course, collecting 1000 labels per example to estimate soft labels becomes prohibitively expensive very quickly. 
However it may not be necessary to collect that many labels in practice.
To determine how many labels are needed to arrive at a reasonable estimate of the soft label distributions, we randomly sampled crowd workers from our dataset one at a time.
At each step, we used the sampled workers responses to estimate the soft labels for each example and calculated the Kullback Liebler divergence (KL-Divergence) between the true soft label distributions and the sampled soft label distributions:$D_{KL}(p \vert \vert q) = - \sum_i P(i) \log \frac{P(i)}{Q(i)}$, where $P$ is the true soft label distribution estimated from the full data set and $Q$ is the sampled soft label distribution.
Figure \ref{fig:kld} plots the KL-Divergence averaged over the number of data set examples (180) as a function of the number of crowd workers selected.\footnote{We truncate the x-axis to focus on the lower values.} 
We plot results for 5 runs of the random sampling procedure.
As the figure shows, the average KL-Divergence approaches 0 well before all 1000 labels are necessary.

When sampling randomly, the average difference drops very quickly, and is very low with as few as 15 or 20 labels per example.
Active learning techniques could reduce this number further, either by selecting ``good annotators'' or identifying examples for which more labels are needed.
This is left for future work.

To confirm the observation that significantly fewer labels are necessary, we randomly sampled 20 annotators from the dataset, used their responses to estimate the soft label distributions, and re-trained the LSTM model with SLMG-I using CCE as the soft label loss function.
We ran this training 10 times, where each time we sampled a new selection of 20 annotators for estimating the soft label distributions.
The average accuracy for these models was $76.9$ and the standard deviation was $0.3$.
These models perform as well as the model using the distributions learned from 1000 annotators, with significantly less annotation cost.

\section{Related Work}
\label{sec:background}

Other work on modeling uncertainty in labels is Knowledge Distillation~\cite{hinton2015distilling}.  
In Knowledge Distillation, output probabilities of a complex expert model are used as input to a simpler model so the simpler model can learn to generalize based on the output weights of the expert model.
%The expectation is that how an expert model assigns output weights can be used to reduce overfitting in the simpler model.
A key distinction between Knowledge Distillation and our work is that the expert model that is distilling its knowledge was still trained with a single class label as the gold standard, and the expert passes its uncertainty to the simpler model.
In our work we capture uncertainty at the original training data, in order to induce generalization as part of the original training.
%In addition, previous work has only been applied to image data, not language data.
%Also Kajino work \cite{kajino2012convex}.
%Deep learning with noisy labels \cite{reed2015training, sukhbaatar2015training, lee2013pseudo}.

This work is related to the idea of ``crowd truth'' and collecting and using annotations from the crowd~\cite{kajino2012convex,inel2014crowdtruth}.
We use the CrowdTruth assumption that disagreement between annotators provides signal about data ambiguity and should be used in the learning process.
%CrowdTruth includes several metrics to calculate likelihoods of different events with regards to particular examples and particular annotators. 
%In those cases, particularly with regards to annotators, the metrics are used to identify potential low-quality annotators for removal.
%In this work we have a large number of annotations for each example (1000 annotations per example), and therefore we assume that any issues of annotator quality will be ``drowned out'' by the large number of annotations.
%Therefore we do not need to identify and remove annotations, and instead can use raw annotation metrics instead of the CrowdTruth metrics.
In addition this work is closely related to the idea of Label Distribution Learning (LDL) from Computer Vision (CV)~\cite{geng2016label}.
For training and testing, LDL assumes that $y$ is a probability distribution over labels.
With LDL, the goal is to learn a distribution over labels.
However in our case we would still like to learn a classifier that outputs a single class, while using the distribution over training labels as a measure of uncertainty in the data.
We use the distribution over labels to represent the uncertainty associated with different examples in order to improve model training.

%\begin{comment}
There are several other areas of study regarding how best to use training data that are related to this work.
Re-weighting or re-ordering training examples is a well-studied and related area of supervised learning.
Often examples are re-weighted according to some notion of difficulty, or model uncertainty~\cite{bengio_curriculum_2009,chang2017active}.
In particular, the internal uncertainty of the model is used as the basis for selecting how training examples are weighted.
However, model uncertainty is dependent upon the original training data the model was trained on, while here we use an external human measure of uncertainty.
Curriculum learning (CL) is a training procedure where models are trained to learn simple concepts before more complex concepts are introduced~\cite{bengio_curriculum_2009}.
CL training for neural networks can improve generalization and speed up convergence.
They demonstrate the effectiveness of curriculum learning on several tasks and draw a comparison with boosting and active learning~\cite{freund_decision-theoretic_1997}.
Our representation of uncertainty via soft labels can be thought of as a measure of difficulty (i.e. more uncertainty is associated with more difficult examples).

Finally, this work is related to transfer learning and domain adaptation~\cite{caruana1995learning,bengio2011deep,bengio2012deep}, but with an important distinction.
Transfer learning and domain adaptation repurpose representations learned for a source domain to facilitate learning in a target domain.
In this paper we want to improve performance in the source domain by fine-tuning with data from the source domain with distributions over class labels.
This work differs from domain adaptation and transfer learning in that we are not adding data from a different domain or applying a learned model to a new task.
Instead, we are augmenting a single classification task by using a richer representation of where the data lies within the class labels to inform training.
The goal is that by fine tuning with a distribution over labels, a model will be less likely to overfit on a training set.
%\end{comment}
%Additional related work includes curriculum learning, active learning, and transfer learning \cite{caruana1995learning,bengio_curriculum_2009,bengio2011deep,bengio2012deep,Druck:2011:TIT:2063576.2063712,yosinski_2014_transfer,chang2017active}.
To the best of our knowledge this is the first work to use a subset of soft labeled data for fine-tuning, whereas previous work used an all-or-none approach (all hard or soft labels).

\section{Discussion}
\label{sec:discussion}

In this paper we have introduced SLMG, a fine-tuning approach to training that can improve classification performance by leveraging uncertainty in data.
In the NLI task, incorporating the more informative class distribution labels leads to improved performance under certain training setups.
By introducing specialized supplemental data the model is able to update its representations to boost performance.
With SLMG, a learning model can update parameters according to a gold-standard that allows for uncertainty in predictions, as opposed to the classic case where each training example should be equally important during parameter updates.
Training examples with higher degrees of uncertainty within a human population have less of an effect on gradient updates than those examples where confidence in the label is very high as measured by the crowd.

SLMG is an easy fix, but it is not a silver bullet for improving generalization.
In our experiments we found that under different training settings SLMG can improve performance for the different models.
It is worthwhile to experiment with SLMG to see if and how it can improve performance on other NLP tasks.
NLI is a particularly good use case for SLMG because of the ambiguity inherent in language and the potential disagreements that can arise from different interpretations of text.
In addition, further experimentation with the way soft labels are generated can lead to further generalization improvements.

There are limitations to this work.
One bottleneck is the requirement for having a large amount of human labels for a small number of examples, which goes against the traditional strategy for crowdsourcing label-generation. 
However one can probably estimate a reasonable distribution over labels with significantly fewer labels than obtained here for each example (Figure \ref{fig:kld}).
%On the other hand, the new SA dataset of human response can be used for modeling parameters such as difficulty as in \cite{lalor2016beyond}.
Identifying a suitable number using active learning techniques is left for future work.

While SLMG requires soft labels, it does not necessarily require \textit{human-annotated} soft labels. Rather, SLMG only requires some measure of uncertainty between training examples as part of the generalization step.
This can come from human annotators, an ensemble of machine learning models, or some other pre-defined uncertainty metric.
In our experiments we demonstrate the validity of SLMG using an existing data set from which we can extract soft labels, and leave experimentation with different soft label generation methods to future work.

Future work includes investigation into data sets that can be used with SLMG and why certain fine-tuning sets lead to better performance in certain scenarios.
Experiments with different loss functions (e.g. KL-Divergence) and different data can help to understand how SLMG affects the representations learned by a model. 
Our results suggest that future work training DNNs to learn a distribution over labels can lead to further improvements.

%\newpage
\fontsize{9.0pt}{10.0pt} \selectfont
\bibliographystyle{aaai}
\bibliography{jlalor}

\begin{thebibliography}{}

\bibitem[\protect\citeauthoryear{Aroyo and Welty}{2015}]{aroyo2015truth}
Aroyo, L., and Welty, C.
\newblock 2015.
\newblock Truth is a lie: Crowd truth and the seven myths of human annotation.
\newblock {\em AI Magazine} 36(1):15--24.

\bibitem[\protect\citeauthoryear{Bachrach \bgroup et al\mbox.\egroup
  }{2012}]{ICML2012Bachrach_597}
Bachrach, Y.; Graepel, T.; Minka, T.; and Guiver, J.
\newblock 2012.
\newblock How to grade a test without knowing the answers --- a bayesian
  graphical model for adaptive crowdsourcing and aptitude testing.
\newblock In {\em Proceedings of the 29th International Conference on Machine
  Learning},  1183--1190.
\newblock New York, NY, USA: Omnipress.

\bibitem[\protect\citeauthoryear{Bengio \bgroup et al\mbox.\egroup
  }{2009}]{bengio_curriculum_2009}
Bengio, Y.; Louradour, J.; Collobert, R.; and Weston, J.
\newblock 2009.
\newblock Curriculum learning.
\newblock In {\em Proceedings of the 26th International Conference on Machine
  Learning},  41--48.
\newblock ACM.

\bibitem[\protect\citeauthoryear{Bengio \bgroup et al\mbox.\egroup
  }{2011}]{bengio2011deep}
Bengio, Y.; Bastien, F.; Bergeron, A.; Boulanger-lew, N.; Breuel, T.;
  Chherawala, Y.; Cisse, M.; Côté, M.; Erhan, D.; Eustache, J.; Glorot, X.;
  Muller, X.; Lebeuf, S.~P.; Pascanu, R.; Rifai, S.; Savard, F.; and Sicard, G.
\newblock 2011.
\newblock Deep learners benefit more from out-of-distribution examples.
\newblock In {\em AISTATS},  164--172.

\bibitem[\protect\citeauthoryear{Bengio}{2012}]{bengio2012deep}
Bengio, Y.
\newblock 2012.
\newblock Deep learning of representations for unsupervised and transfer
  learning.
\newblock {\em ICML Unsupervised and Transfer Learning} 27:17--36.

\bibitem[\protect\citeauthoryear{Bowman \bgroup et al\mbox.\egroup
  }{2015}]{bowman_large_2015}
Bowman, S.~R.; Angeli, G.; Potts, C.; and Manning, D.~C.
\newblock 2015.
\newblock A large annotated corpus for learning natural language inference.
\newblock In {\em Proceedings of the 2015 Conference on Empirical Methods in
  Natural Language Processing},  632--642.
\newblock Association for Computational Linguistics.

\bibitem[\protect\citeauthoryear{Caruana}{1995}]{caruana1995learning}
Caruana, R.
\newblock 1995.
\newblock Learning many related tasks at the same time with backpropagation.
\newblock {\em Advances in Neural Information Processing Systems}  657--664.

\bibitem[\protect\citeauthoryear{Chang, Learned-Miller, and
  McCallum}{2017}]{chang2017active}
Chang, H.-S.; Learned-Miller, E.; and McCallum, A.
\newblock 2017.
\newblock Active bias: Training a more accurate neural network by emphasizing
  high variance samples.
\newblock In {\em Advances in Neural Information Processing Systems}.

\bibitem[\protect\citeauthoryear{Chen \bgroup et al\mbox.\egroup
  }{2017}]{Chen-Qian:2017:ACL}
Chen, Q.; Zhu, X.; Ling, Z.; Wei, S.; Jiang, H.; and Inkpen, D.
\newblock 2017.
\newblock Enhanced lstm for natural language inference.
\newblock In {\em Proceedings of the 55th Annual Meeting of the Association for
  Computational Linguistics (ACL 2017)}.
\newblock Vancouver: ACL.

\bibitem[\protect\citeauthoryear{Dagan, Glickman, and
  Magnini}{2006}]{dagan_pascal_2006}
Dagan, I.; Glickman, O.; and Magnini, B.
\newblock 2006.
\newblock The {PASCAL} {Recognising} {Textual} {Entailment} {Challenge}.
\newblock In {\em Machine {Learning} {Challenges}. {Evaluating} {Predictive}
  {Uncertainty}, {Visual} {Object} {Classification}, and {Recognising}
  {Tectual} {Entailment}}. Springer.
\newblock  177--190.

\bibitem[\protect\citeauthoryear{Dawid and Skene}{1979}]{dawid1979maximum}
Dawid, A.~P., and Skene, A.~M.
\newblock 1979.
\newblock Maximum likelihood estimation of observer error-rates using the em
  algorithm.
\newblock {\em Journal of the Royal Statistical Society. Series C (Applied
  Statistics)} 28(1):20--28.

\bibitem[\protect\citeauthoryear{Druck and
  McCallum}{2011}]{Druck:2011:TIT:2063576.2063712}
Druck, G., and McCallum, A.
\newblock 2011.
\newblock Toward interactive training and evaluation.
\newblock In {\em Proceedings of the 20th ACM International Conference on
  Information and Knowledge Management}, CIKM '11,  947--956.
\newblock New York, NY, USA: ACM.

\bibitem[\protect\citeauthoryear{Freund and
  Schapire}{1997}]{freund_decision-theoretic_1997}
Freund, Y., and Schapire, R.~E.
\newblock 1997.
\newblock A {Decision}-{Theoretic} {Generalization} of {On}-{Line} {Learning}
  and an {Application} to {Boosting}.
\newblock {\em J. Comput. Syst. Sci.} 55(1):119--139.

\bibitem[\protect\citeauthoryear{Geng}{2016}]{geng2016label}
Geng, X.
\newblock 2016.
\newblock Label distribution learning.
\newblock {\em IEEE Transactions on Knowledge and Data Engineering}
  28(7):1734--1748.

\bibitem[\protect\citeauthoryear{Gururangan \bgroup et al\mbox.\egroup
  }{2018}]{gururangan2018annotation}
Gururangan, S.; Swayamdipta, S.; Levy, O.; Schwartz, R.; Bowman, S.; and Smith,
  N.~A.
\newblock 2018.
\newblock Annotation artifacts in natural language inference data.
\newblock In {\em Proceedings of the 2018 Conference of the North American
  Chapter of the Association for Computational Linguistics: Human Language
  Technologies, Volume 2 (Short Papers)}, volume~2,  107--112.

\bibitem[\protect\citeauthoryear{Halevy, Norvig, and
  Pereira}{2009}]{halevy2009unreasonable}
Halevy, A.; Norvig, P.; and Pereira, F.
\newblock 2009.
\newblock The unreasonable effectiveness of data.
\newblock {\em IEEE Intelligent Systems} 24(2):8--12.

\bibitem[\protect\citeauthoryear{Hinton, Vinyals, and
  Dean}{2015}]{hinton2015distilling}
Hinton, G.; Vinyals, O.; and Dean, J.
\newblock 2015.
\newblock Distilling the knowledge in a neural network.
\newblock {\em arXiv preprint arXiv:1503.02531}.

\bibitem[\protect\citeauthoryear{Hochreiter and
  Schmidhuber}{1997}]{hochreiter_long_1997}
Hochreiter, S., and Schmidhuber, J.
\newblock 1997.
\newblock Long {Short}-{Term} {Memory}.
\newblock {\em Neural Computation} 9(8):1735--1780.

\bibitem[\protect\citeauthoryear{Inel and Aroyo}{2017}]{inel2017harnessing}
Inel, O., and Aroyo, L.
\newblock 2017.
\newblock Harnessing diversity in crowds and machines for better ner
  performance.
\newblock In {\em European Semantic Web Conference},  289--304.
\newblock Springer.

\bibitem[\protect\citeauthoryear{Inel \bgroup et al\mbox.\egroup
  }{2014}]{inel2014crowdtruth}
Inel, O.; Khamkham, K.; Cristea, T.; Dumitrache, A.; Rutjes, A.; van~der Ploeg,
  J.; Romaszko, L.; Aroyo, L.; and Sips, R.-J.
\newblock 2014.
\newblock Crowdtruth: Machine-human computation framework for harnessing
  disagreement in gathering annotated data.
\newblock In {\em International Semantic Web Conference},  486--504.
\newblock Springer.

\bibitem[\protect\citeauthoryear{Kajino, Tsuboi, and
  Kashima}{2012}]{kajino2012convex}
Kajino, H.; Tsuboi, Y.; and Kashima, H.
\newblock 2012.
\newblock A convex formulation for learning from crowds.
\newblock In {\em Twenty-Sixth AAAI Conference on Artificial Intelligence}.

\bibitem[\protect\citeauthoryear{Kamar, Kapoor, and
  Horvitz}{2015}]{kamar2015identifying}
Kamar, E.; Kapoor, A.; and Horvitz, E.
\newblock 2015.
\newblock Identifying and accounting for task-dependent bias in crowdsourcing.
\newblock In {\em Third AAAI Conference on Human Computation and
  Crowdsourcing}.

\bibitem[\protect\citeauthoryear{Lalor \bgroup et al\mbox.\egroup
  }{2018}]{lalor2018irt}
Lalor, J.~P.; Wu, H.; Munkhdalai, T.; and Yu, H.
\newblock 2018.
\newblock Understanding deep learning performance through an examination of
  test set difficulty: A psychometric case study.
\newblock In {\em Proceedings of the 2018 Conference on Empirical Methods in
  Natural Language Processing}.
\newblock Association for Computational Linguistics.

\bibitem[\protect\citeauthoryear{Lalor, Wu, and Yu}{2016}]{lalor2016beyond}
Lalor, J.~P.; Wu, H.; and Yu, H.
\newblock 2016.
\newblock Building an evaluation scale using item response theory.
\newblock In {\em Proceedings of the 2016 Conference on Empirical Methods in
  Natural Language Processing},  648--657.
\newblock Association for Computational Linguistics.

\bibitem[\protect\citeauthoryear{Landis and Koch}{1977}]{landis1977measurement}
Landis, J.~R., and Koch, G.~G.
\newblock 1977.
\newblock The measurement of observer agreement for categorical data.
\newblock {\em biometrics}  159--174.

\bibitem[\protect\citeauthoryear{Munkhdalai and
  Yu}{2017}]{munkhdalai2016neural}
Munkhdalai, T., and Yu, H.
\newblock 2017.
\newblock Neural semantic encoders.
\newblock In {\em Proceedings of the 15th Conference of the European Chapter of
  the Association for Computational Linguistics}.
\newblock Association for Computational Linguistics.

\bibitem[\protect\citeauthoryear{Neubig \bgroup et al\mbox.\egroup
  }{2017}]{dynet}
Neubig, G.; Dyer, C.; Goldberg, Y.; Matthews, A.; Ammar, W.; Anastasopoulos,
  A.; Ballesteros, M.; Chiang, D.; Clothiaux, D.; Cohn, T.; Duh, K.; Faruqui,
  M.; Gan, C.; Garrette, D.; Ji, Y.; Kong, L.; Kuncoro, A.; Kumar, G.;
  Malaviya, C.; Michel, P.; Oda, Y.; Richardson, M.; Saphra, N.; Swayamdipta,
  S.; and Yin, P.
\newblock 2017.
\newblock Dynet: The dynamic neural network toolkit.
\newblock {\em arXiv preprint arXiv:1701.03980}.

\bibitem[\protect\citeauthoryear{Pennington, Socher, and
  Manning}{2014}]{pennington2014glove}
Pennington, J.; Socher, R.; and Manning, C.~D.
\newblock 2014.
\newblock Glove: Global vectors for word representation.
\newblock In {\em Empirical Methods in Natural Language Processing (EMNLP)},
  1532--1543.

\bibitem[\protect\citeauthoryear{Poliak \bgroup et al\mbox.\egroup
  }{2018}]{poliak2018hypothesis}
Poliak, A.; Naradowsky, J.; Haldar, A.; Rudinger, R.; and Van~Durme, B.
\newblock 2018.
\newblock Hypothesis only baselines in natural language inference.
\newblock In {\em Proceedings of the Seventh Joint Conference on Lexical and
  Computational Semantics},  180--191.

\bibitem[\protect\citeauthoryear{{Theano Development
  Team}}{2016}]{2016arXiv160502688short}
{Theano Development Team}.
\newblock 2016.
\newblock {Theano: A {Python} framework for fast computation of mathematical
  expressions}.
\newblock {\em arXiv e-prints} abs/1605.02688.

\bibitem[\protect\citeauthoryear{Tokui \bgroup et al\mbox.\egroup
  }{2015}]{chainer_learningsys2015}
Tokui, S.; Oono, K.; Hido, S.; and Clayton, J.
\newblock 2015.
\newblock Chainer: a next-generation open source framework for deep learning.
\newblock In {\em Proceedings of Workshop on Machine Learning Systems
  (LearningSys) in The Twenty-ninth Annual Conference on Neural Information
  Processing Systems (NIPS)}.

\bibitem[\protect\citeauthoryear{Williams, Nangia, and
  Bowman}{}]{williams_multinli_2018}
Williams, A.; Nangia, N.; and Bowman, S.~R.
\newblock A broad-coverage challenge corpus for sentence understanding through
  inference.
\newblock In {\em Proceedings of the 2018 Conference of the North American
  Chapter of the Association for Computational Linguistics: Human Language
  Technologies (NAACL)}.

\end{thebibliography}

\end{document}